\documentclass{article}
\PassOptionsToPackage{numbers,square,sort&compress}{natbib}
\usepackage[preprint]{neurips_2025}
\makeatletter
\renewcommand{\@noticestring}{Preprint.}
\makeatother
\usepackage[utf8]{inputenc}
\usepackage[T1]{fontenc}
\usepackage{hyperref}
\usepackage{url}
\usepackage{booktabs}
\usepackage{amsmath,amssymb}
\usepackage{microtype}
\usepackage{graphicx}
\usepackage{xcolor}

\newcommand{\KL}{\mathrm{KL}}
\newcommand{\bpw}{\,bpw}

\title{NariQ: Sub-3-Bit Quantization for Long-Horizon Agentic Tasks}

\author{%
  Doyeob (Toby) Kim\\
  Nari Labs\\
  \texttt{toby@narilabs.com}
}

\begin{document}
\maketitle

\begin{abstract}
Long-horizon agentic tasks require a model to call tools over many steps at contexts of 100K tokens or more, yet
many low-bit quantization methods are calibrated on short web text and judged on perplexity. This mismatch is costly: existing
quantizations of Qwen3.8-27B at or below 3 bits per weight match BF16 on shorter tasks like AIME, yet in published results
solve at most 72\% as many tasks as BF16 on DeepSWE, a long-horizon agentic benchmark. We propose NariQ, a
trellis-coded post-training quantization method that calibrates on tokens generated by the BF16 model in agent sessions. On
Qwen3.8-27B, NariQ at 3.0 and 2.5 bits per weight stays level with BF16 on two long-horizon agentic benchmarks,
DeepSWE and Terminal Bench, as well as on AIME and RealWorldQA. Our checkpoint at 2.0 bits per weight trails this performance. To our knowledge, these are the first
quantizations at or below 3 bits per weight shown to match their full-precision version on long-horizon
agentic tasks. We release the files as
\href{https://huggingface.co/nari-labs/Qwen3.8-27B-NariQ-GGUF}{\texttt{Qwen3.8-27B-NariQ-GGUF}}, with inference available
through our llama.cpp fork at \url{https://github.com/nari-labs/nari-llama.cpp}.
\end{abstract}

\section{Introduction}

Language models are increasingly used as agents working on long-horizon tasks. For example, a single task in
DeepSWE~\cite{deepswe}, a popular software engineering benchmark, takes an agent on average about 150 tool calls and
50 minutes, with a median context of 92K tokens.

Near 4 bits per weight, quantization already preserves such agentic performance. NVIDIA's NVFP4 version of Qwen3.8-27B~\cite{qwen38} stays
within 1.6 points of BF16 on Terminal Bench~\cite{nvfp4qwen} and solves at least as many DeepSWE tasks as
BF16~\cite{imecquant}.

Below 4 bits, existing quantized models still do well on short tasks, but not on long-horizon ones. For example, the
3-bit GSQ-RCO quantization of Qwen3.8-27B~\cite{gsqrco} stays within a point of BF16 on AIME (96.9\% against 97.7\%).
Yet in Unsloth's reported results it fully solves only 63\% as many DeepSWE tasks as BF16 (Table~\ref{tab:main}).

We therefore ask whether we can quantize a model to 3 bits per weight or fewer without losing its long-horizon agent
capability. The answer decides whether capable agents can run on consumer GPUs, where the weights must fit into
8--32\,GB.

A better weight format alone is unlikely to close this gap. Random rotations make weights close to
Gaussian~\cite{quip,quipsharp}, and on Gaussians, trellis codes are already within about 10\% of the rate-distortion
bound at 2 bits (mean squared error 0.069 against 0.063)~\cite{qtip}. We instead ask which tokens a quantized agent
must get right. Many low-bit methods are calibrated on short chunks of web text~\cite{gptq} and judged on perplexity
and short benchmarks, which say little about agents. They weight every calibration token equally, whereas an agent's
quality depends on the tokens it writes.

We introduce NariQ, a post-training quantization recipe for long-horizon agentic use at 2.0, 2.5 and 3.0 bits per
weight. NariQ keeps the standard trellis code and changes which tokens the loss counts, which modules get which
widths, and what data the scales are tuned on. It sets each module's width from an estimate of
that module's own loss, which stays informative for trellis codes. It codes the output layer in bands by how likely
the model is to predict each token, keeping every band at 2 bits or more to protect rare tokens.

We evaluate agentic work, math, visual understanding and multilingual tasks. NariQ at 3.0 and
2.5\bpw{} is level with BF16 within noise on all benchmarks we ran; at 2.0\bpw, it trails BF16.

\section{Method}
\label{sec:method}

A file at $B$\bpw{} uses $B$ bits per text-model parameter, including codes, scales, labels, embeddings and the
output layer. To build a file $\hat\theta$ whose next-token distribution stays close to BF16's, we (1) quantize modules
at several integer widths, (2) estimate their loss from gradients, (3) allocate widths within the budget, and
(4) tune the scales.

\begin{figure}[t]
  \centering
  \includegraphics[width=0.75\linewidth]{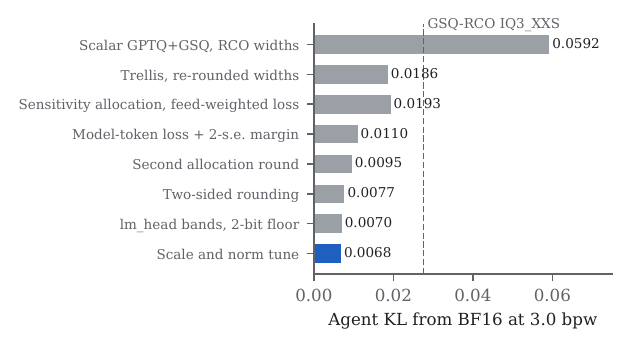}
  \caption{Development sequence at 3.0\bpw{}, ending at an earlier checkpoint. Trellis coding has 3.2$\times$ lower
  agent KL than GPTQ+GSQ, with the scalar search's probabilities re-rounded under trellis costs.
  Subsequent changes reduce KL a further 2.7$\times$.
  The model-token step also adds the two-standard-error margin, so it does not isolate masking alone. Dashed: GSQ-RCO IQ3\_XXS.}
  \label{fig:ladder}
\end{figure}

We trellis-quantize each rotated matrix~\cite{qtip}, using input- and output-side statistics to compensate rounding
error as in YAQA~\cite{yaqa}. We retain input-only rounding unless two-sided rounding lowers held-out
curvature-weighted error by at least 5\%; almost every module uses both sides. NariQ adds four changes;
Figure~\ref{fig:ladder} shows their development sequence at 3.0\bpw.

First, the end-to-end loss counts only model-written tokens. Agent traces contain long tool outputs and file
contents that can dominate an all-token loss. For a calibration trace $x$ (Section~\ref{sec:data}), let $\mathcal{M}$ index the tokens generated
by BF16. Our objective is
\begin{equation}
\mathcal{L}=\frac{1}{|\mathcal{M}|}\sum_{t\in\mathcal{M}}\KL\big(p_{\text{BF16}}(\cdot\mid x_{<t})\,\big\|\,p_{\hat\theta}(\cdot\mid x_{<t})\big).
\end{equation}
Both models receive the same prefix $x_{<t}$, including tool outputs. The mask selects prediction targets;
it does not remove context tokens.

Second, widths are allocated by each module's own loss term. Since the KL is minimized at BF16, we use a second-order
expansion around its weights and neglect interactions between modules' errors, so the total loss is a sum of
per-module terms. Take one module
and flatten its weights into a vector: $W$ at BF16 and $W_k=W+E_k$ at width $k$. With $H$ the module's Hessian
block at BF16, its loss term at width $k$ is $\ell_k=\frac12\langle E_k,HE_k\rangle$. We take gradients
of $\mathcal{L}$ on a file with every module at some width. Let $c$ be the module's width in that file and $G$ its
weight gradient; under the approximation, $G\approx HE_c$.
RCO~\cite{rco} scores a move to another width $k\neq c$ by the first-order change
\begin{equation}
\langle G, W_k-W_c\rangle \approx \langle E_c,HE_k\rangle-\langle E_c,HE_c\rangle .
\end{equation}
The first term measures how much the errors at the two widths align. They barely do: in held-out curvature
estimates for four probe matrices, the curvature-weighted cosine between the 3-bit error and the 2-, 4- and 5-bit
errors is 0.035, 0.009 and 0.004. We therefore neglect this term. The score is then about $-2\ell_c$ for every
$k\neq c$, whereas the quadratic loss change is $\ell_k-\ell_c$, so the score misses the loss at the destination
width.

NariQ instead uses the score to estimate the current loss: averaging $\frac12\langle G,W_c-W_k\rangle$ over
$k\neq c$ gives an estimate of $\ell_c$. The gradient estimates $\ell_c$ only at the
current width. To predict the other widths, we assume a module's loss scales with the code's error: with $r_k$ the
trellis code's measured mean squared error at width $k$, which falls about 4$\times$ per added bit, we predict
$\ell_k\approx\ell_c\,r_k/r_c$. For example, a module's loss at 2 bits is predicted as about 3.9 times its loss at
3 bits.

Scaling a noisy gradient estimate to a distant width amplifies its noise. We combine estimates from several width
assignments, rescaling to a common width and weighting by inverse variance. We use the mean plus two standard
errors and apply a small positive floor.

\begin{samepage}
Let $\hat\ell_{m,k}$ be the resulting predicted loss for module $m$ at width $k$. The allocation objective is
\begin{equation}
\min_{k_m\in\mathcal{K}_m}\sum_m\hat\ell_{m,k_m}
\quad\text{subject to}\quad C_{\mathrm{fixed}}+\sum_m n_m k_m\le BN,
\end{equation}
where $n_m$ is the module's parameter count, $\mathcal{K}_m$ its allowed widths, $N$ the total text-model parameter
count, and $C_{\mathrm{fixed}}$ the width-independent bits for scales, labels and tensors outside the allocation.
\end{samepage}
Choosing one width per module makes this a
multiple-choice knapsack problem. We use dynamic programming over a discretized budget, favoring the fullest
feasible allocation. An exact-cost check enforces the storage limit, but global optimality is not guaranteed. The loss estimate fails at 1 bit: at 2.0\bpw,
moving the 12 modules the allocator had put at 1 bit up to 2 bits lowered KL by 7.2\% for only 0.021\bpw, far more
than predicted. Every linear layer therefore stays at 2 bits or more.

Third, the output layer gets more bits for likely tokens. Only about 13\% of its 248,320 tokens appear as generated
tokens in our calibration traces, although every row contributes to the softmax. For a token with BF16 probability
$p$, the KL curvature with respect to its logit is $p(1-p)$, which is close to $p$ for all but the most likely
tokens. We therefore rank rows by their mean predicted probability, as a proxy for sensitivity, and sort them into 16 bands of about 15,500 rows. We code each band at
its own width, as a separate module in the allocation. Rare tokens still need some precision: in our tests, bands at 1 bit lowered KL on most traces but raised it on
traces that use rare tokens, such as multilingual text, so every band keeps at least 2 bits. At 3.0\bpw{} the bands lower
the mean per-trial KL by 12.1\%. At 2.0\bpw{} the input embedding is also banded. It has no predicted
probability, so its rows are ranked by how often each token appears in the input, and unlike output bands, its
rare-token bands may use 1 bit.

Fourth, scale tuning uses general data as well as coding traces. With codes and widths fixed, we tune every coded
matrix's scales and every norm weight to minimize $\mathcal{L}$ on coding agent traces and BF16 replies to general prompts (math, science,
chat, code and long documents), weighted 70:30. An earlier
2.0\bpw{} file tuned on coding traces alone scored 77.3\% on AIME against BF16's 97.7\%; adding the general data raised it to 86.0\%.

\textbf{Format.} We release the files as NariQ GGUFs~\cite{llamacpp}. The format rotates every matrix with randomized Hadamard
blocks~\cite{quipsharp} of power-of-two size. In attention and DeltaNet layers, one side of most matrices is first
shuffled by a seeded permutation. Because every block has power-of-two size, the format needs no Hadamard sizes tied
to a particular model. The codebook is computed from the trellis state, as in EXL3~\cite{exl3}, so kernels decode
the codes inside the matrix multiply.

\section{Evaluating Low-Bit Files on Agentic Tasks}
\label{sec:eval}

\begin{table}[t]
\centering
\small
\caption{Agent benchmarks, in our harness and as Unsloth reports them from its own harness
(Section~\ref{sec:baselines}). Solved: DeepSWE tasks fully solved, of 113. Score: Terminal Bench mean test reward. \% BF16: share
of the same harness's BF16. Bold: best quantized file in each harness.}
\label{tab:main}
\setlength{\tabcolsep}{5pt}
\begin{tabular}{@{}lr rr rr@{}}
\toprule
 & & \multicolumn{2}{c}{DeepSWE v1.1} & \multicolumn{2}{c}{Terminal Bench 2.1} \\
\cmidrule(lr){3-4}\cmidrule(l){5-6}
File & bpw & Solved & \% BF16 & Score & \% BF16 \\
\midrule
\multicolumn{6}{@{}l}{\emph{Our harness}} \\
BF16 & 16 & 44 & 100 & 62.6 & 100 \\
Ours 3.0 & 3.00 & \textbf{47} & \textbf{107} & \textbf{63.8} & \textbf{102} \\
Ours 2.5 & 2.50 & 46 & 105 & 60.2 & 96 \\
Ours 2.0 & 2.00 & 33 & 75 & 52.4 & 84 \\
\midrule
\multicolumn{6}{@{}l}{\emph{Unsloth's harness (reported)}} \\
BF16 & 16 & 45.7 & 100 & 59.15 & 100 \\
Unsloth UD-IQ4\_XS & 4.16 & 41.0 & 90 & 55.7 & 94 \\
Unsloth UD-Q3\_K\_XL & 3.80 & \textbf{42.5} & \textbf{93} & 56.7 & 96 \\
EXL3 3.00\bpw & 3.77 & 35.3 & 77 & 55.3 & 93 \\
GSQ-RCO IQ3\_S & 3.50 & 40.4 & 88 & 54.9 & 93 \\
Unsloth UD-IQ3\_S & 3.47 & 39.0 & 85 & \textbf{59.8} & \textbf{101} \\
GSQ-RCO IQ3\_XXS & 3.00 & 29.0 & 63 & 51.2 & 87 \\
EXL3 2.20\bpw & 2.91 & 11.0 & 24 & 43.1 & 73 \\
Unsloth UD-Q2\_K\_XL & 2.85 & 33.0 & 72 & 53.0 & 90 \\
GSQ-RCO IQ2\_S & 2.75 & 28.0 & 61 & 42.7 & 72 \\
Unsloth UD-IQ2\_S & 2.50 & 24.5 & 54 & 44.6 & 75 \\
GSQ-RCO IQ2\_XS & 2.50 & 19.0 & 42 & 37.8 & 64 \\
Unsloth UD-IQ2\_XS & 2.27 & 14.5 & 32 & 41.0 & 69 \\
\bottomrule
\end{tabular}
\end{table}

\subsection{Data and Benchmarks}
\label{sec:data}

\textbf{Calibration and development traces.} Agent sessions of the BF16 model on open-source software tasks, split
by repository into calibration, development and reserve sets, with every repository shared with DeepSWE removed.

\textbf{Offline KL sets.} KL from BF16 at model-written tokens on held-out agent traces (agent KL), on a reserve set
never used for selection, and on general and math replies (Appendix~\ref{app:kl}).

\textbf{DeepSWE v1.1}~\cite{deepswe}, at task-repository commit \texttt{0b9fabb}. A software engineering benchmark of
113 tasks from 91 repositories in five languages. Claude Code drives the model, which SGLang~\cite{sglang} serves.
The agent runs with thinking on at maximum effort, up to 65,536 output tokens per request and a limit of 3 hours
per task. Each task gets one attempt; agent failures count as zero, and no task is rerun. We report the DeepSWE score, the
mean fraction of a task's new tests that pass, the number of tasks solved, where every new and existing test passes,
and paired differences with standard errors.

\textbf{Terminal Bench 2.1}~\cite{terminalbench}. Command-line tasks, each with its own tests and time limit. We run
the 82 tasks Unsloth reports on, with 3 attempts each, the same agent and settings as DeepSWE, and each task's own
time limit and internet access. We report the mean test reward over all attempts; a trial left without a grade counts
as zero.

\textbf{AIME 2025+2026~\cite{aime}, RealWorldQA~\cite{realworldqa} and MGSM~\cite{mgsm}.} Math, visual understanding and multilingual tasks, to measure what a file keeps
outside coding. Thinking is on at the chat template's default effort, with up to 131,072 output tokens for AIME and
32,768 for the others; a reply cut at the limit counts as wrong. We sample 8 replies per AIME problem (avg@8), 3 per RealWorldQA question and 1 per MGSM question, as in simple-evals~\cite{simpleevals}.

\textbf{Settings shared by every benchmark.} Every file, BF16 included, runs with the same settings: a 262,144-token context window, the model card's thinking-mode sampling (temperature 1.0, top-p
0.95, top-k 20), and no retries.

\subsection{Baselines}
\label{sec:baselines}

\textbf{Agent benchmarks (Table~\ref{tab:main}).} We run BF16 and our files in our harness. For other quantized
files we use the DeepSWE and Terminal Bench 2.1 results Unsloth publishes for this model~\cite{unsloth}: its Dynamic
V3.1 GGUFs, GSQ-RCO's GGUFs~\cite{gsqrco} and the EXL3 files~\cite{exl3}. Unsloth uses mini-swe-agent~\cite{minisweagent} and at least
three passes per DeepSWE task~\cite{unsloth31}; we use Claude Code and one attempt. Table~\ref{tab:main} reports each
file relative to its own harness's BF16, but this normalization does not remove harness differences. Unsloth's
solved counts are its reported rates times 113, read from its chart to about $\pm0.5$ tasks.

\textbf{Offline KL (Appendix~\ref{app:kl}).} We decode every published file of this model near our budgets, from
GSQ-RCO, DASLab~\cite{gsq}, Unsloth and EXL3, exactly to BF16 and score it with our code.

\section{Results}
\label{sec:results}

\subsection{Agent benchmarks}

At 3.0 and 2.5\bpw, our files are level with BF16 on DeepSWE (Table~\ref{tab:main}). They solve 47 and 46 tasks
against BF16's 44, and their DeepSWE scores are 84.9\% and 84.5\% against BF16's 83.0\% (paired by task, $+1.9$ and
$+1.5$ points, s.e.\ 3.2 and 3.6). At 2.0\bpw, our file solves 33 tasks and scores 75.7\%, 7.3 points under BF16
(s.e.\ 3.3). In Unsloth's reported results, every file at or below 3.0\bpw{} solves at most 72\% as many tasks as
its BF16.

On Terminal Bench 2.1, our 3.0 and 2.5\bpw{} files score 102\% and 96\%
of our BF16, within noise of it (paired by task, $+1.2$ and $-2.4$ points, s.e.\ 3.5 and 3.2), and our 2.0\bpw{} file scores
84\%, 10.2 points under (s.e.\ 3.5). The reported files at or below 3.0\bpw{} score 64--90\% of
Unsloth's BF16.

\subsection{Beyond coding}

\begin{table}[t]
\centering
\small
\caption{Benchmarks outside coding (\%).}
\label{tab:side}
\begin{tabular}{@{}lrrr@{}}
\toprule
File & AIME & RealWorldQA & MGSM \\
\midrule
BF16 & 97.7 & 82.4 & 94.0 \\
Ours 3.0 & 98.3 & 82.4 & 93.1 \\
Ours 2.5 & 96.5 & 81.3 & 91.8 \\
Ours 2.0 & 86.0 & 80.9 & 90.6 \\
\bottomrule
\end{tabular}
\end{table}

General ability holds (Table~\ref{tab:side}): on AIME, RealWorldQA and MGSM, our 3.0\bpw{} file is within a point of
BF16 and our 2.5\bpw{} file within 2.2 points. At 2.0\bpw, the largest drop is on AIME, 11.7 points.

\subsection{Size and speed}

NariQ's text-only GGUFs, including embeddings and output layers, occupy 10.1, 8.4 and 6.8\,GB at 3.0, 2.5 and
2.0\bpw. Vision adds a shared 0.9\,GB BF16 projector; the original 55.6\,GB BF16 checkpoint includes vision and MTP.
At these three budgets, our llama.cpp fork on one RTX PRO 6000 generates 89, 99 and 112 tokens/s at a 4K-token
context, and 71, 77 and 85 at 64K, for a single session.

\section{Related Works}

\textbf{Codes and rounding.} Low-bit formats include scalar and grouped codes~\cite{gptq,awq}, and vector and
trellis codes~\cite{quipsharp,aqlm,qtip}, often applied after random
rotations~\cite{quip,quipsharp}. EXL3~\cite{exl3} ships the trellis family for local inference. GPTQ~\cite{gptq} and
LDLQ~\cite{quip} round with the input Hessian, and YAQA~\cite{yaqa} adds the output side. We use QTIP's bitshift
trellis with a computed codebook.

\textbf{Bit allocation and recovery.} Mixed-precision allocation by sensitivity is established~\cite{hawqv3}, and
search methods such as EvoPress~\cite{evopress} and RCO~\cite{rco} optimize widths against the end-to-end loss;
GSQ-RCO~\cite{gsqrco} publishes files of this model built that way. Variable bit-width quantization~\cite{vbq} learns
widths per group, down to about 1 bit on the output layer. Fine-tuning scales and other continuous parameters after
quantization is common~\cite{qtip,efficientqat}. Our allocator differs in reading the end-to-end gradient as each
module's own loss term, which remains informative for codes whose errors at different widths are uncorrelated.

\textbf{Generated calibration data.} FAQ~\cite{faq} regenerates calibration data with a larger model from the same
family. OnQ~\cite{onq} recovers low-bit reasoning through distillation on the quantized student's own trajectories.
NariQ instead uses fixed BF16-generated tool-use traces, restricts its loss to model-written positions, and allocates
bits against that objective. We evaluate the resulting files on long-horizon agentic tasks.

\section{Limitations and Future Work}

While our 3.0 and 2.5\bpw{} files are level with BF16 on all benchmarks we ran, evaluations on more models, more agent
harnesses and more agentic benchmarks remain future work. At 2.0\bpw, our file trails BF16 by 7.3 points on DeepSWE
and 11.7 on AIME, and we hypothesize that training beyond scales and norms can recover it.

The multi-token prediction module is not yet quantized, and our llama.cpp fork is tuned for one GPU generation
(SM120); other GPUs run it slower, and GPUs older than Ampere fall back to the CPU. We also plan to quantize the KV
cache and activations, so long contexts fit on 16\,GB consumer GPUs, and extend the
study to more models and more budgets at or below 3 bits.

\section{Conclusion}

NariQ calibrates and allocates bits around model-written tokens in agent sessions. On Qwen3.8-27B, its 3.0 and
2.5\bpw{} files are level with BF16 on long-horizon agentic benchmarks. We release the files as
\href{https://huggingface.co/nari-labs/Qwen3.8-27B-NariQ-GGUF}{\texttt{Qwen3.8-27B-NariQ-GGUF}}, with inference available
through our llama.cpp fork at \url{https://github.com/nari-labs/nari-llama.cpp}.

\paragraph{Acknowledgments.} AI tools assisted with writing, literature search, implementation, and research analysis.

{\footnotesize
\setlength{\bibsep}{0.5pt}
\bibliographystyle{abbrvnat}
\bibliography{refs}
}

\appendix

\section{Offline KL against public releases}
\label{app:kl}

In Table~\ref{tab:rivals}, at 2.5 and 3.0\bpw{} the best published file in each budget group has
1.7--4.9$\times$ our KL on every set, with the same or more bits. At 2.0\bpw, Unsloth's UD-IQ2\_XS, with 0.27\bpw{} more, is slightly ahead on the math set (0.0819 against 0.0829); we lead on the other sets, by 2.1$\times$ on agent KL.

\begin{table}[!h]
\centering
\footnotesize
\renewcommand{\arraystretch}{0.92}
\caption{KL from BF16 at model-written tokens (lower is better). Agent and Reserve: held-out coding-agent sessions
(Reserve never used for any choice). General: math, science, chat, code and long-document replies. Math: held-out math
replies. Best per class in bold.}
\label{tab:rivals}
\begin{tabular}{@{}lrrrrr@{}}
\toprule
File & bpw & Agent & Reserve & General & Math \\
\midrule
Ours 2.0 & 2.00 & \textbf{0.0284} & \textbf{0.0270} & \textbf{0.0841} & 0.0829 \\
Unsloth UD-IQ1\_M & 2.00 & 0.1823 & 0.2071 & 0.2931 & 0.3105 \\
Unsloth UD-IQ2\_XXS & 2.14 & 0.0732 & 0.0851 & 0.1115 & 0.1024 \\
EXL3 SC\_1.40 & 2.18 & 0.3135 & 0.3614 & 0.5536 & 0.5809 \\
Unsloth UD-IQ2\_XS & 2.27 & 0.0593 & 0.0686 & 0.0899 & \textbf{0.0819} \\
EXL3 SC\_1.60 & 2.36 & 0.2415 & 0.2813 & 0.4181 & 0.4310 \\
\midrule
Ours 2.5 & 2.50 & \textbf{0.0117} & \textbf{0.0109} & \textbf{0.0360} & \textbf{0.0314} \\
GSQ-RCO IQ2\_XS & 2.50 & 0.0623 & 0.0763 & 0.0928 & 0.0742 \\
Unsloth UD-IQ2\_S & 2.50 & 0.0453 & 0.0527 & 0.0680 & 0.0618 \\
EXL3 SC\_1.80 & 2.55 & 0.1611 & 0.1893 & 0.2830 & 0.2873 \\
EXL3 SC\_2.00 & 2.73 & 0.0573 & 0.0660 & 0.1060 & 0.1012 \\
GSQ-RCO IQ2\_S & 2.75 & 0.0435 & 0.0530 & 0.0613 & 0.0523 \\
\midrule
Unsloth UD-Q2\_K\_XL (V3.1) & 2.85 & 0.0275 & 0.0318 & 0.0407 & 0.0364 \\
EXL3 SC\_2.20 & 2.91 & 0.0484 & 0.0562 & 0.0836 & 0.0823 \\
Ours 3.0 & 3.00 & \textbf{0.0063} & \textbf{0.0060} & \textbf{0.0173} & \textbf{0.0148} \\
GSQ-RCO IQ3\_XXS & 3.00 & 0.0276 & 0.0333 & 0.0376 & 0.0308 \\
Unsloth UD-IQ3\_XXS & 3.14 & 0.0239 & 0.0279 & 0.0390 & 0.0384 \\
DASLab GSQ 3-bit & 3.24 & 0.0362 & 0.0436 & 0.0391 & 0.0283 \\
\bottomrule
\end{tabular}
\end{table}

\end{document}